\documentclass[letterpaper, 10 pt, conference]{ieeeconf}  %

\IEEEoverridecommandlockouts                              %

\usepackage[english]{babel}

\usepackage{graphicx}
\usepackage{animate}
\usepackage{epsfig} 				%
\usepackage{epstopdf}

\usepackage{listings}
\usepackage{color}
\usepackage{nameref}
\usepackage{hyperref}
\usepackage{amsmath}	 			%
\usepackage{amssymb}  				%

\usepackage{dsfont}			%
\usepackage{mathtools}

\usepackage{epigraph}
\usepackage{lscape}
\usepackage[]{nomencl}				%
\usepackage{algorithm}
\usepackage{algorithmic}
\usepackage{multicol}
\usepackage{multirow}
\usepackage{etoolbox}

\usepackage{caption}
\usepackage{subcaption}
\usepackage{wrapfig}

\usepackage{siunitx}

\usepackage{floatflt}

\usepackage{url}

\let\labelindent\relax
\usepackage{enumitem}
\usepackage{longtable}
\usepackage{booktabs}

\newcommand{\email}[1]{\href{mailto:#1}{\nolinkurl{#1}}}

\newcommand{\fig}[1]{Fig.~\ref{#1}}

\def\vzero{{\bm{0}}}

\def\vtheta{{\bm{\theta}}}
\def\va{{\bm{a}}}

\def\vp{{\bm{p}}}

\def\vr{{\bm{r}}}

\def\vv{{\bm{v}}}

\def\vtheta{{\boldsymbol{\theta}}}

\def\mI{{\bm{I}}}

\def\mO{{\bm{O}}}

\def\mT{{\bm{T}}}

\DeclareMathAlphabet{\mathsfit}{\encodingdefault}{\sfdefault}{m}{sl}
\SetMathAlphabet{\mathsfit}{bold}{\encodingdefault}{\sfdefault}{bx}{n}

\def\gD{{\mathcal{D}}}

\def\gL{{\mathcal{L}}}

\def\gN{{\mathcal{N}}}

\def\RR{\mathbb{R}}

\usepackage{bm}
\usepackage{comment}
\usepackage{xcolor} %
\newcommand{\citet}[1]{\cite{#1}}
\newcommand{\citep}[1]{\cite{#1}}

\usepackage[font=small]{caption}
\let\ACMmaketitle=\maketitle
\renewcommand{\maketitle}{\begingroup\let\footnote=\thanks \ACMmaketitle\endgroup}

\makeatletter
\newcommand*\titleheader[1]{\begingroup\gdef\@titleheader{#1}\let\footnote=\thanks\endgroup}
\AtBeginDocument{%
  \let\st@red@title\@title
  \def\@title{%
  \begin{flushleft}
    \vspace{-2.15em}
    \bgroup\normalfont\small\@titleheader\par\egroup
    \vspace{-16pt}\par\noindent\rule{\textwidth}{0.1pt}
    \end{flushleft}
    \vskip1.0em\st@red@title
        }

}

\title{\LARGE \bf {On the Importance of Tactile Sensing for Imitation Learning: \\ A Case Study on Robotic Match Lighting}}

\author{Niklas Funk$^{1}$, Changqi Chen$^{1}$, Tim Schneider$^{1}$, Georgia Chalvatzaki$^{1}$, Roberto Calandra$^{2}$, Jan Peters$^{1}$%
\thanks{$^{1}$Technical University of Darmstadt, Darmstadt, Germany}
\thanks{$^{2}$LASR Lab, TU Dresden, Dresden, Germany}
\thanks{Corresponding author: Niklas Funk. Email: niklas@robot-learning.de}
\thanks{\copyright 2026 IEEE. Personal use of this material is permitted.
Permission from IEEE must be obtained for all other uses, in any current or future media, including reprinting/republishing this material for advertising or promotional purposes, creating new collective works, for resale or redistribution to servers or lists, or reuse of any copyrighted component of this work in other works. DOI: 10.1109/LRA.2026.3677716}%
}

\titleheader{\textit{Preprint}}

\begin{document}

\maketitle
\thispagestyle{empty}
\pagestyle{empty}

\begin{abstract}
	The field of robotic manipulation has advanced significantly in recent years.
At the sensing level, several novel tactile sensors have been developed, capable of providing accurate contact information.
On a methodological level, learning from demonstrations has proven an efficient paradigm to obtain performant robotic manipulation policies.
The combination of both holds the promise to extract crucial contact-related information from the demonstration data and actively exploit it during policy rollouts.
However, this integration has so far been underexplored, most notably in dynamic, contact-rich manipulation tasks where precision and reactivity are essential.
This work therefore proposes a multimodal, visuotactile imitation learning framework that integrates a modular transformer architecture with a flow-based generative model, enabling efficient learning of fast and dexterous manipulation policies.
We evaluate our framework on the dynamic, contact-rich task of robotic match lighting - a task in which tactile feedback influences human manipulation performance.
The experimental results highlight the effectiveness of our approach and show that adding tactile information improves policy performance, thereby underlining their combined potential for learning dynamic manipulation from few demonstrations.
Project website: \url{https://sites.google.com/view/tactile-il}.

\end{abstract}

\IEEEpeerreviewmaketitle

\section{INTRODUCTION}

Robotic manipulation still remains far from matching human dexterity and efficiency~\cite{sampath2023review, kroemer2021review}.
A promising direction toward closing this gap is leveraging human demonstration data for learning robotic manipulation through imitation \cite{ravichandar2020recent, zhaoaloha, chi2023diffusion}, thereby actively exploiting humans' task understanding and advanced manipulation capabilities.
Although numerous studies have shown that access to touch sensing benefits human manipulation performance \cite{vallbo1984properties, edin1992independent, pavlova2015activity}, the majority of current works in imitation learning for manipulation are still missing out on this modality \cite{zhaoaloha, chi2023diffusion, funk2024actionflow}.
This raises the question of \textit{whether and how learning robotic manipulation policies from human demonstrations could also benefit from incorporating tactile sensing}.

\begin{figure}
\begin{center}
\includegraphics[width=0.85\columnwidth]{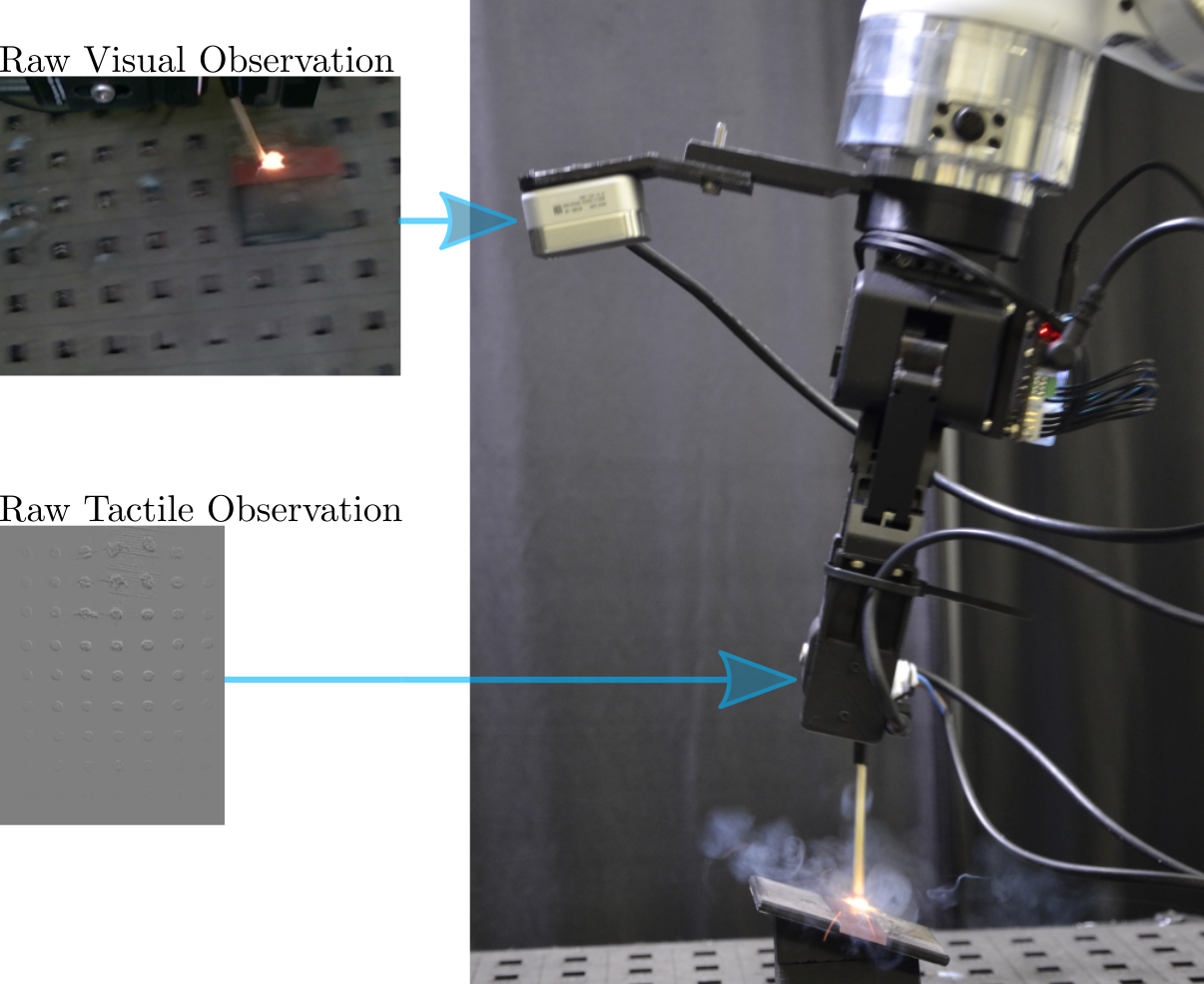}
\end{center}
\caption{
Autonomous rollout of a policy that is conditioned on visual and tactile observations illustrated on the left.
The policy controls the robot and, thereby, the contact configuration between the match and striker paper.
The policy ensures sufficient force and velocity, resulting in successfully igniting the match.
This work highlights the importance of tactile sensing for reliably solving the dynamic and delicate task of lighting up matches.}
\label{fig:visuotactile_rollout}
\vspace{-0.45cm}
\end{figure}

This work approaches this question by studying the impact of touch sensing for learning a dynamic task, namely, igniting matches.
We argue that match lighting is a challenging and effective testbed because the task requires both dynamic motion and compliance~\cite{kronander_match}, thereby extending established manipulation benchmarking tasks such as pick-and-place or insertion, which have recently seen significant progress through multisensory learning approaches~\cite{dong2021tactile, hansen2022visuotactile, huang20243d, yu2023mimictouch, li2023see, feng2025play}.
Moreover, it is a task for which there is evidence that the availability of touch sensing impacts human performance~\cite{JohannsonYouTube}.
Despite the task's relevance, to the best of our knowledge, it has only previously been investigated by Kronander et al.~\cite{kronander_match}, who considered fixed match grasp poses in a precisely calibrated setup without high-dimensional observations.
Our work instead addresses more complicated scenarios, including varying grasp poses and striker paper orientations, while solely considering local embodied sensing, i.e., RGB wrist-camera images, the end effector velocity, and the information from an event-based optical tactile sensor (cf. Fig. \ref{fig:visuotactile_rollout}).

To address the intricate challenges of this dynamic manipulation task, we also introduce a multimodal learning from demonstrations framework.
The proposed framework is built upon an expressive multimodal flow matching policy \cite{chen2023riemannian} with a modular and efficient transformer-based policy architecture. This combination enables reactivity through fast, real-time inference and the comparison of different encoding and training strategies.
To minimize human effort, we prioritize data efficiency by relying only on 20 demonstrations in most experiments.
Our experiments demonstrate that the proposed framework achieves robust match lighting, while consistently outperforming competitive baselines.
The performance benefits also generalize to another task, namely, whiteboard erasing, confirming its effectiveness for diverse contact-rich tasks.
The results reveal that success is driven by both visuotactile integration and our proposed approach. Furthermore, vision-only policies can benefit from a masked training procedure that leverages tactile data exclusively during training.

Overall, we present a multimodal framework for efficiently learning robust and reliable manipulation policies suitable for dynamic tasks such as lighting matches.
Moreover, we present a masked training procedure that exploits the tactile signals only during training and allows for increased success rates of vision-only policies.
Lastly, we contribute an extensive evaluation conducted in our modular real-world match-lighting testing environment, spanning more than 600 rollouts.
Experiments across different policies and configurations provide crucial insights regarding the importance of incorporating tactile sensing into policy learning and demonstrate the competitive performance of our proposed approach and its components.

\section{RELATED WORK}
\label{sec:related}

The emergence of commercial \cite{yuan2017gelsight} and open-source tactile sensors \cite{ward2018tactip, evetac_paper} has drawn increased attention to advancing robotic manipulation by integrating touch sensing~\cite{li2020review}.

One approach to obtain tactile manipulation policies is reinforcement learning \cite{dong2021tactile,hansen2022visuotactile, church2022tactile}.
Due to the high sample complexity, previous works either rely on fast simulation \cite{hansen2022visuotactile, church2022tactile, Bi2021ZeroShotST, ojaghi2025curriculum} or on carefully designed hardware environments~\cite{dong2021tactile} enabling extensive real-world autonomous exploration.
Since our task of match lighting is challenging to simulate, and safety considerations hinder real-world autonomous exploration, this work investigates efficiently learning match lighting policies from few real-world expert demonstrations, thereby significantly reducing the data requirements.

Learning robotic manipulation policies from demonstration data \cite{ravichandar2020recent} has lately received increasing attention \cite{zhaoaloha, chi2023diffusion, funk2024actionflow, ablett2021seeing, mandlekar2021matters}.
Several works achieved advanced real-world manipulation by training generative models on expert demonstrations and utilizing improved data-collection devices~\cite{zhaoaloha, chi2023diffusion, Zhao-RSS-ACT, mukashev2024bts}.
However, most works focus on pick-and-place manipulation and only incorporate RGB or RGB-D cameras as external sensors, without considering tactile information \cite{zhaoaloha, chi2023diffusion, funk2024actionflow, Zhao-RSS-ACT}.
This work follows the current efforts and proposes an efficient and modular multi-modal framework for learning from demonstrations by leveraging a generative model trained as a policy.
Yet, it differs in that it considers tactile sensors as input modality and investigates the contact-rich and dynamic manipulation task of igniting matches.
Only recently, a few works investigated adding tactile sensing capabilities into imitation learning frameworks~\cite{huang20243d, li2023see, ablett2023multimodal}, moving toward tasks with more intricate contact configurations, such as wiping or tight insertion \cite{yu2023mimictouch, liu2025maniwav,xue2025reactive}.
While \cite{huang20243d,liu2025maniwav} leverages diffusion policy for policy learning and \cite{li2023see, ablett2023multimodal} use a standard mean-squared error behavioural cloning loss, none of the works investigated a flow-matching-based policy, which we find to be a key component for high success rates in the dynamic match ignition task.
Concurrent efforts to enhance policy reactivity include the Reactive Diffusion Policy (RDP)~\cite{xue2025reactive}, which employs a hierarchical decomposition and separately trains a slow latent diffusion policy and a fast, touch-conditioned tokenizer, and Implicit RDP~\cite{chen2025implicitrdp}, which proposes an approach to fuse multi-frequency visual and tactile observations within a unified model.
While \cite{xue2025reactive,chen2025implicitrdp} explicitly exploit the temporal interplay between lower- and higher-frequency observational signals, our work explores a complementary direction. 
We focus on developing a modular and effective policy architecture that, together with the flow-matching objective, enables reactive and performant control for dynamic tasks such as match lighting.
Furthermore, we introduce a masked training procedure, showcasing that considering tactile observations during training can enhance the inference performance of vision-only policies.

From a task-level perspective, \cite{kronander_match} is closest related as it also investigates learning match lighting policies from human demonstrations.
To achieve good task success rates, they propose a varying stiffness controller learned through information from a human-robot interface.
Instead of learning a variable stiffness controller, this work directly learns a reactive policy capable of controlling the contact forces by varying the desired target poses.
Moreover, this work extends \cite{kronander_match} by considering a more realistic experimental setup, including varying match poses, striker paper orientations, and conditioning the policies on high-dimensional image and tactile observations.

Overall, we contribute a framework for learning visuotactile robotic match lighting policies from human demonstrations and showcasing that tactile sensing is crucial for learning high performance policies on dynamic tasks.

\section{LEARNING MATCH LIGHTING POLICIES FROM DEMONSTRATIONS}
\label{sec:approach}

	This section describes our approach for learning the dynamic manipulation skill of match lighting from few real-world demonstrations.
This work exclusively considers local, embodied sensing, i.e., images from a wrist-mounted camera, an open-source Evetac \cite{evetac_paper} tactile sensor, and local velocity information (cf. Fig.~\ref{fig:inference_illustration}).
The following sections detail the learning framework, the policy architecture, the data collection, and the policy inference procedure.

\subsection{Fast and Reactive Multimodal Policies through Conditional Flow Matching}

Our multimodal policy learning framework leverages a generative model as policy.
Given the current observations, the model should output actions close to the demonstrations.
Since the match lighting task is delicate and requires reactivity, we propose to learn a policy using flow matching~\cite{black2024pi_0}.
We learn an SE(3)-Rectified Linear flow model~\cite{funk2024actionflow} that generates high-quality samples within low inference times.
We impose a flow in SE(3), as the model should output the desired future trajectory of the robot end-effector, a sequence of $N=16$ SE(3) poses, $\mT_a = (T^1_a,\dots, T^N_a) \in SE(3)^N$.

The idea in conditional rectified linear flow matching is to impose a straight line path between samples from a noise distribution $\va_{t=0}{\sim}\gN(\vzero,\mI)$ and samples from the dataset $\va_{t=1}{\in}\gD$. The intermediate waypoints of the flow are thus defined by $\va_{t}{=}t\va_{t=1}+(1{-}t)\va_{t=0}$, $t\in[0,1]$.
The objective is then to learn the velocity field of this path $\mathrm{d}\va_{t}/\mathrm{d}t$, such that during inference, samples can be generated by starting from a random initial sample and iteratively refining it through Euler integration given a learned estimate of the velocity field.
For our case of SE(3) action poses that should be generated $\va_{t} {=} (\vp_t {\in} \RR^3,\vr_t {\in} SO(3))$, we decouple the translational \& rotational flow components and obtain $\dot{\vp}_t {=} (\vp_1 {-} \vp_t)/(1{-}t)$ \& $\dot{\vr}_t = (\text{Log}(\vr_t^{-1}\vr_1))/(1{-}t)$ for the velocity field.
Given the training data, we then train a parameterized flow matching model $\vv_{\vtheta}(\vp_t, \vr_t,\mO, t)$, that, conditioned on the current observation $\mO$ and ``action" pose, outputs translation and rotation velocities (${\vv_{\vtheta,\vp}\in\RR^3}$ \& ${\vv_{\vtheta,\vr}\in\RR^3}$).
The model is trained by minimizing $\gL = || \vv_{\vtheta,\vp} {-} \dot{\vp}_t ||^2 + || \vv_{\vtheta,\vr} {-} \dot{\vr}_t ||^2$.
During inference, we sample actions by iteratively refining random initial actions through $\vp_{k+1} {=} \vp_k {+} \vv_{\vtheta}(\vp_t, \vr_t,\mO, t)\Delta t$ \& $\vr_{k+1} {=} \vr_k \text{Exp}(\Delta t \vv_{\vtheta}(\vp_t, \vr_t,\mO, t))$.
Note that we define the flow for the entire action sequence of 16 poses.

\subsection{Policy Architecture}
\label{sec:policy_arch}

\begin{figure}
\begin{center}
\includegraphics[width=0.8\columnwidth]{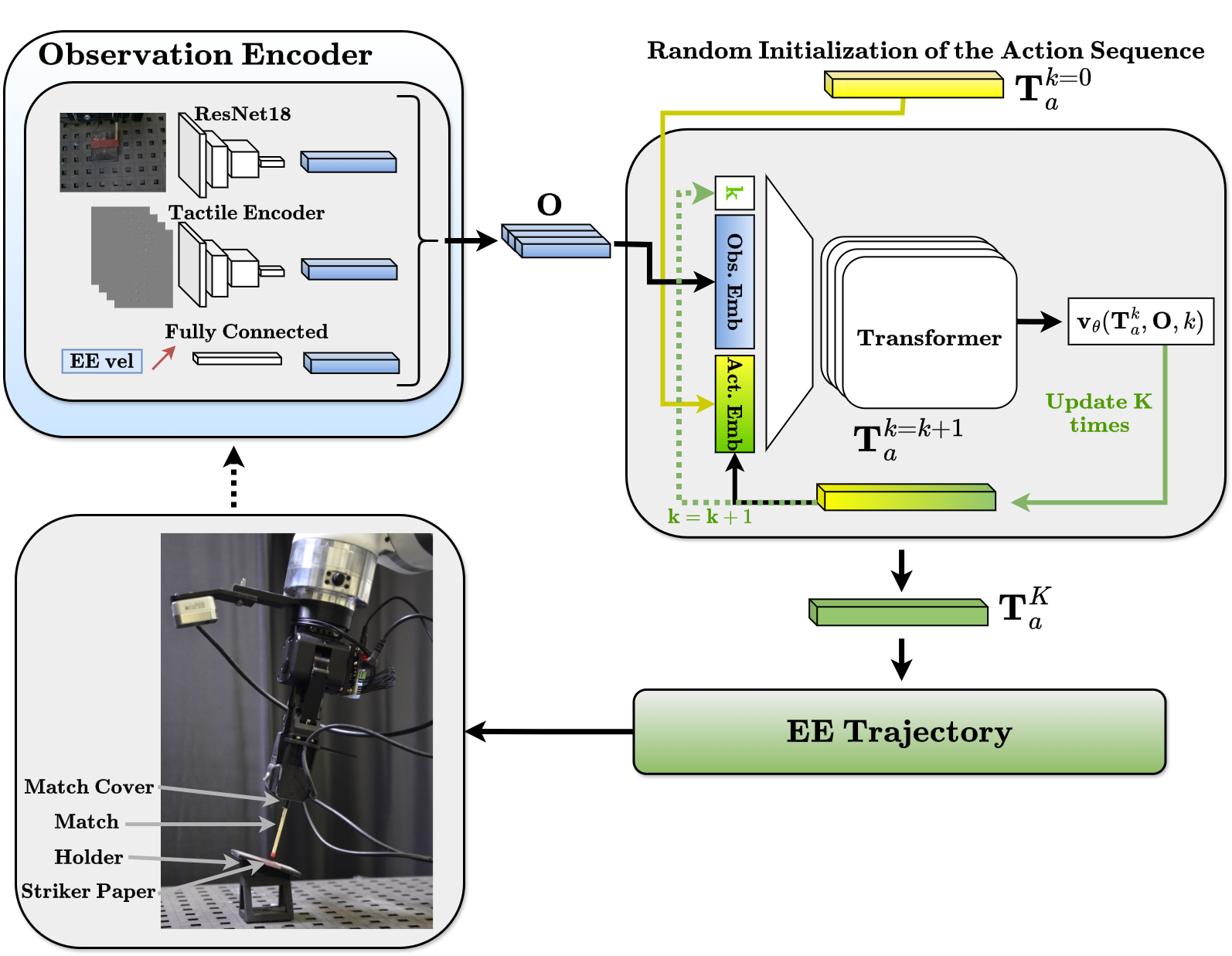}
\end{center}
\caption{
Method Overview.
The observations are first encoded individually inside the observation encoder and brought into a common shape, i.e., each modality contributes a latent vector of a fixed shape.
These latent vectors, together with the current action sequence \& time index, are then passed into the transformer, which outputs velocities to iteratively refine the action sequence through flow matching.
Once the final desired end-effector trajectory is retrieved, it is sent to the robot and tracked using a Cartesian Impedance Controller.
Note that we only apply the first action to maintain reactivity.
}
\label{fig:inference_illustration}
\vspace{-0.45cm}
\end{figure}

Building on the previous section, we employ a parameterized SE(3)-rectified linear flow matching model for action generation.
At the core of this policy is a multimodal transformer architecture that receives observations from multiple sensors, including the RGB camera image, the current end-effector velocity, and, when available, Evetac's tactile readings.
Transformers are particularly suitable for this task as they can seamlessly handle multiple multimodal observations \cite{xu2023multimodal}.
The resulting policy architecture is illustrated in \fig{fig:inference_illustration}.

Observations provide the critical information for action refinement.
To facilitate comparing sensor combinations, we ensure modularity by first encoding each modality individually into a 64-dimensional latent vector.
The first 5 entries of this 64-dimensional vector are learnable weights that should inform the transformer about the type of observation modality.
These latent vectors then serve as the input to a 4-layer, 4-head transformer for refining the action sequence.
Importantly, the latent observations and entries of the action sequence enter the transformer as individual tokens.
This tokenized design facilitates seamless evaluation across sensor combinations and enables a masked training procedure, where modalities are stochastically omitted to regularize the policy.
The image observations are processed through a pre-trained ResNet 18 \cite{he2016deep} or by training the ResNet from scratch.
For the tactile observations, we consider the pre-trained model from \cite{evetac_paper}, and training this architecture from scratch.
Inside the transformer, the inputs exchange information with each other and update their embeddings through multi-head attention.
In its standard implementation, all inputs exchange information with each other (including self-connections).
Herein, we configure the transformer's attention mask to be fully connected across observation tokens, whereas action tokens solely cross-attend to observation tokens.
The value of the action tokens thus does not influence the update of the observation tokens.
This choice is made because only the observations contain information about how to update the action sequence, whereas the action sequence contains only noise, particularly at the beginning.
Moreover, the self-attention within the action tokens is configured to attend only to previous actions.
In addition to this action-masking scheme, in the experiments, we also investigate the effectiveness of a masked training procedure.
During masked training, we employ stochastic observation masking, in which tactile inputs are omitted with 50\% probability.
Due to this stochasticity in the input, the policy has to better align the latent representations of vision and touch so that it can generate good outputs in both cases, i.e., when touch is available and when it is not.

The transformer's final output is the updated action features representing the velocity vectors for the iterative refinement, which is repeated $K{=5}$ times.
After obtaining the final action sequence, it is sent to the controller for execution on the robot.
Using this generative model as policy yields online action generation (cf.~\fig{fig:inference_illustration}).

\subsection{Data Collection}

As in~\cite {kronander_match}, demonstrations are collected through kinesthetic teaching.
This ensures that the human demonstrator directly feels the interaction between match and striker paper, and has been crucial for high success rates during data collection.
From a task-level perspective, to light up the match, the match tip must first be brought into contact with the striker paper.
Subsequently, the match tip has to be moved along the striker paper while applying sufficient force with sufficient velocity.

Figs.~\ref{fig:visuotactile_rollout}~\&~\ref{fig:inference_illustration} depict the components of our real-world match lighting environment.
Throughout the demonstrations, we record all sensor data, i.e., images from the wrist-mounted Intel RealSense D405 camera, an Evetac \cite{evetac_paper} tactile sensor mounted within a Robotis RH-P12-RN gripper that is attached to a 7-DoF Franka Panda, and the local end-effector velocity.
We also record the end-effector poses traversed by the robot. They contain the trajectory information that the robot should follow.
Yet, we emphasize that the policy framework only receives local poses expressed in the current end-effector frame.
While Evetac naturally returns asynchronous event information, for compatibility with the other sensors, we convert the events into image form by integrating them for a duration of \SI{40}{\ms}.
We also collect all the other sensor information at \SI{25}{\Hz}.
Since the task is delicate, image (or tactile image) resolution might be crucial. Thus, we maintain a high resolution of $320 \times 240$ pixels.
As shown in Figs.~\ref{fig:visuotactile_rollout}~\&~\ref{fig:inference_illustration}, for the wrist camera images, we ensure that the match and its tip are always fully observable.
Moreover, we found that using the striking surfaces of regular paper matchboxes yielded limited durability, and therefore 3D-printed a thin rectangular plate to hold the striker paper.
In its standard configuration, the plate is raised and placed with an angle of $20^{\circ}$ relative to the table (cf. \fig{fig:visuotactile_rollout}).
We used long standard matches with dimensions of \SI[separate-uncertainty = true]{100 \pm 5}{\mm} $\times$ \SI[separate-uncertainty = true]{4 \pm 1}{\mm} $\times$ \SI[separate-uncertainty = true]{4 \pm 1}{\mm} to keep the fire at a sufficient distance from the silicone surfaces of the tactile sensors mounted inside the gripper.
Lastly, we 3D-printed hollow cylindrical cones to cover the upper \SI{45}{\mm} of the matches.
This was necessary to significantly increase the longevity of the silicone gels covering the tactile sensor, which could tear easily when in direct contact with matches.

\subsection{Policy Inference and Robot Control}

We use the Cartesian Impedance Controller from \cite{Nbfigueroafranka_interactive_controllers} for control.
Policy inference is performed asynchronously, and only the first action in the sequence is applied before updating the sequence using the most recent model inference.
The policies run online in real time. Action generation takes only \SI{0.028}{\s} for our largest vision+touch policies on an NVIDIA $3090$.
For more details, please see Appendix A on our website.

\section{EXPERIMENTAL RESULTS}
\label{sec:result}

	\begin{figure}
\begin{center}
\includegraphics[width=0.95\columnwidth]{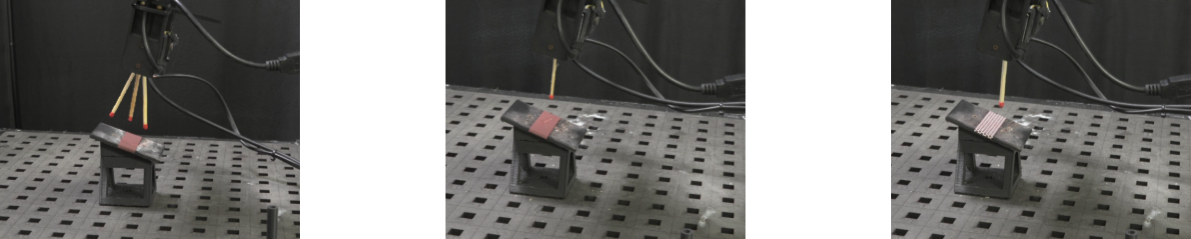}
\end{center}
\caption{
Visualizing the versatility of the initial configurations during the match lighting experiments. Left: Overlay of three distinct initializations in the nominal scenario, highlighting variations in distance and angle between the match and striker paper that need to be handled by the policies.
Middle: Initial configuration for the generalization study utilizing the different, thinner matches.
Right: Experimental setup using the original matches, with a different striker paper to test the policies' robustness.
}
\label{fig:initializations}
\vspace{-0.45cm}
\end{figure}

This section evaluates the effectiveness of our proposed approach and the importance of tactile input. 
It follows five questions:
\textbf{A:} How important is tactile feedback for obtaining performant match lighting policies?
\textbf{B:} Can the vision-only policies benefit from leveraging the tactile information during training? 
\textbf{C:} How does the performance difference evolve when increasing the number of demonstrations?
\textbf{D:} How does our approach perform compared to baselines? and
\textbf{E:} Do the results transfer to the task of whiteboard erasing?

Unless specified otherwise, the evaluations consider robotic match lighting under variable grasping, i.e., varying the grasping location within translational offsets of $\pm$\SI{1}{\cm} \& rotational perturbations of $\pm$\SI{10}{^{\circ}} (cf. Fig. \ref{fig:initializations} - left) while utilizing 20 human demonstrations that have been collected within 1 hour.
All models have been trained for 500 epochs.
We report the mean performance together with the standard deviation across task and model configurations. We trained 3 seeds per combination and evaluated the last checkpoint through 10 rollouts on the real system.
We employ the following naming conventions for the models trained within our framework. 
We refer to the models trained using the standard procedure, where observation modalities remain consistent during both training and inference as "vision+touch", "vision", and "touch".
While the "vision+touch" models leverage both modalities, the "vision" and "touch" policies are conditioned exclusively on visual or tactile inputs, respectively.
Furthermore, vision-only models trained via our proposed stochastic masking procedure (cf. Sec. III-B) are designated as "vision (masked training)".
As such, this procedure is exclusively applied to this model configuration.
Lastly, we note that all models are trained on the same set of demonstrations and are configured to maintain a consistent number of parameters.
Further insights and experiments are provided in the video and appendix (see the supplementary material or our website).

\subsection{How important is tactile feedback for obtaining performant match lighting policies?}
\label{sec:exp_1}

\begin{figure}[t]
\begin{center}
\includegraphics[width=0.95\columnwidth]{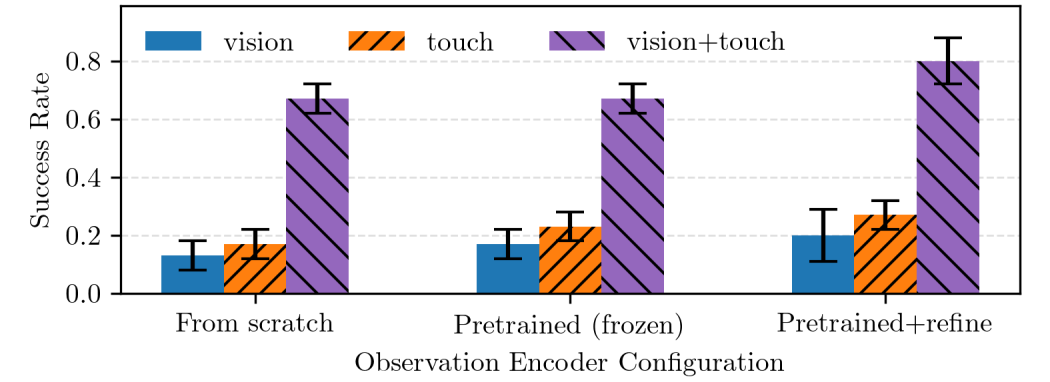}
\end{center}
\caption{Comparing policy success rates (mean \& std dev.) for robotic match lighting.
Apart from different observation modalities, we also consider employing different observation encoding strategies, i.e., training the observation encoders from scratch (left), starting with pretrained encoders and leaving them fixed during training (middle), \& starting with pretrained weights but refining them during training (right).
Across observation encoding strategies, vision+touch policies consistently outperform vision-only policies, underscoring the importance of integrating tactile sensing into the visuomotor policies. Vision+touch policies also outperform a touch-only baseline, underlining that the synergy between visual awareness and reliable tactile perception is crucial for high task success rates.}
\label{fig:success_rates}
\vspace{-0.45cm}
\end{figure}

This experiment compares the performance of vision-only, touch-only, and visuotactile policies under the standard training procedure across different observation encoder configurations.
We train policies with the pre-trained observation encoders (cf.~Fig.~\ref{fig:inference_illustration} \& Sec. III-B) and either freeze ("pretrained (frozen)") or optimize them ("pretrained + refine").
We also investigate training the observation encoders from scratch ("from scratch").
As presented in Fig.~\ref{fig:success_rates}, we observe a significant difference between the vision-only and vision+touch policies in terms of success rate across all observation encoding strategies.
While the best visuotactile policies achieve an average success rate of 80\%, the best performing vision-only policies only reach success rates of up to 20\%.
These findings highlight the effectiveness of incorporating the tactile observations into the vision-only policies, and that visual information alone is insufficient for solving the task reliably.
Moreover, we also find that the visuotactile policies reliably outperform a touch-only baseline.
These improvements are statistically significant (two-sided Fisher’s exact test, $p < 0.01$), with odds ratios ($OR$) of 16.0 against vision-only and 9.3 against touch-only baselines, confirming the performance benefits of the vision+touch policies.
We employed Fisher’s Exact Test given the binary success metrics and the sample size ($N=30$ per group). While all policies were evaluated under the same controlled task distributions and hardware, the test accounts for the inherent stochasticity in grasping poses, striker paper placement, and physical wear across individual trials.

\begin{figure}
\begin{center}
\includegraphics[width=0.95\columnwidth]{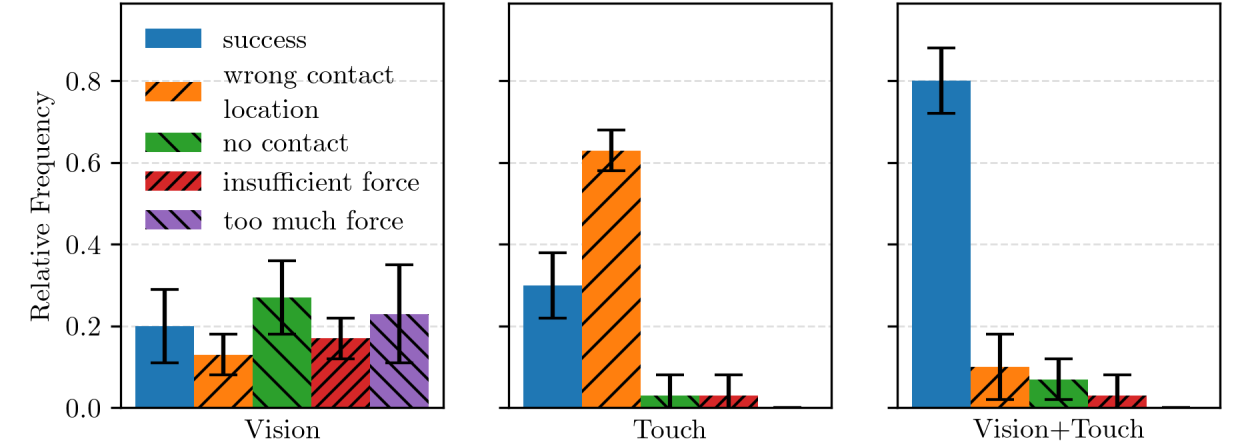}
\end{center}
\caption{Comparing success rates and failure modes (mean \& std dev) for the vision, touch, and vision+touch policies for the pretrained+refine observation encoding strategy (cf. Fig.~\ref{fig:success_rates}).
The vision+touch policies reduce the failure rate by over 40\%, substantially decreasing contact-related failures (not applying force, applying insufficient force, or applying excessive force) compared to vision-only policies, and reduce wrong-contact-location failures compared to touch-only policies.}
\label{fig:failure_analysis}
\vspace{-0.45cm}
\end{figure}

Fig.~\ref{fig:failure_analysis} provides a more detailed comparison by differentiating between failure modes, thereby offering additional insights into the performance discrepancies and the role of the modalities.
We consider four failure types: 1) making contact in the wrong location, i.e., the tip of the match not making contact with the striker paper, 2) not making contact at all, i.e., the policy accelerating along the striker paper without making contact, 3) insufficient contact force, i.e., making contact in the right location but without sufficient force to light up the match, and 4) applying too much force, i.e., the policy pressing the match tip with too much force against the striker paper which results in the match sliding through the fingers.
The last failure case is primarily due to the policy failing to transition from the approaching phase to the acceleration movement along the striker paper.
As shown in Fig.~\ref{fig:failure_analysis}, both the touch-only and the vision-only policies exhibit significantly increased failure rates, albeit for different reasons.
Most failures of the touch-only policies stem from making contact at the wrong location, indicating that the task needs spatial understanding.
Conversely, vision-only policy failures are largely driven by the inability to reliably resolve the current contact state. Specifically, failures characterized by no contact (27\%), insufficient force (17\%), and excessive force (23\%) are most prominent.
In contrast, the vision+touch policies effectively integrate the spatial understanding of the visual modality during the approach phase with tactile features that are effective during the contact-rich interactions. This integration significantly reduces failure rates, thereby underscoring the importance of integrating touch sensing into visuomotor policies. This enables direct perception of the contact-rich interactions, which is a critical capability for reliably solving this dynamic manipulation task.
The few failures of the vision+touch policies are mainly due to making contact in the wrong location (10\%).

\begin{figure}
\begin{center}
\includegraphics[width=0.9\columnwidth]{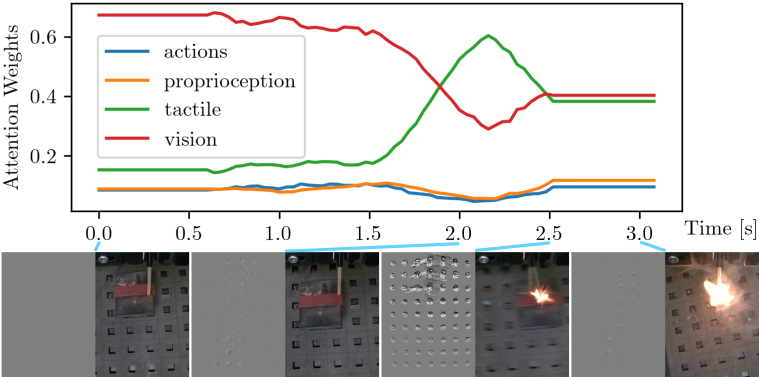}
\end{center}
\caption{Visualizing the evolution of the attention weights over time for a policy rollout.
The bottom images show the task progression.
The plot shows the weights that are attributed to the transformer inputs: 1) the actions, 2) the proprioception observation (end effector velocity), 3) the tactile observations, and 4) the vision observations.
The weights are w.r.t. to updating the fifth action of the desired end-effector trajectory, which is computed for every observation along the rollout.
At the beginning and end of the trajectory (when there are no tactile signals), vision is the most important modality. Once there are changes in contact configuration, touch is the most important modality for action generation, therefore highlighting that touch provides important feedback for controlling the contact configuration.}
\label{fig:attn_analysis}
\vspace{-0.45cm}
\end{figure}

Lastly, Fig. \ref{fig:attn_analysis} illustrates the evolution of the attention weights for the individual transformer inputs w.r.t. updating the fifth action of the sequence for a visuotactile policy (trained using  "pretrained+refine").
It shows how the inputs contribute to the action update.
As shown, initially, the vision input is the most important modality.
This is expected, as the camera information is crucial to moving the robot closer to the striker paper.
The event-based tactile sensor does not provide any information during this phase, as there is no change in contact configuration.
However, once contact is made, the tactile inputs gain importance and become the most important entity.
This holds true until the match ignites, which signals successful task execution.
The other inputs, i.e., attention to other actions and proprioceptive observations, remain low throughout the trajectory.
The evolution of the attention weights, therefore, again underlines the synergy between visual spatial awareness during the approaching movement and the reliable tactile contact perception during the physical interaction.

\subsection{Can the vision-only policies benefit from leveraging tactile information during training?}

While the previous section demonstrated the importance of integrating touch sensing into vision-only policies, this section investigates whether such policies can benefit from using tactile information exclusively during training.
By exploiting the transformer's inherent ability to process variable-length input sequences, we employ a masked training procedure where tactile signals are stochastically omitted with a 50\% probability (cf. Sec.~\ref{sec:policy_arch}).
This forces the model to align the latent spaces to produce consistent outputs regardless of the input observations, thereby biasing the vision encoder to account for the eventual missing tactile information.
We use the pre-trained encoders and optimize them during training because they yield meaningful embeddings from the outset.
We designate the vision-only policies obtained through this method as "vision (masked training)".

\begin{figure}
\begin{center}
\includegraphics[width=0.9\columnwidth]{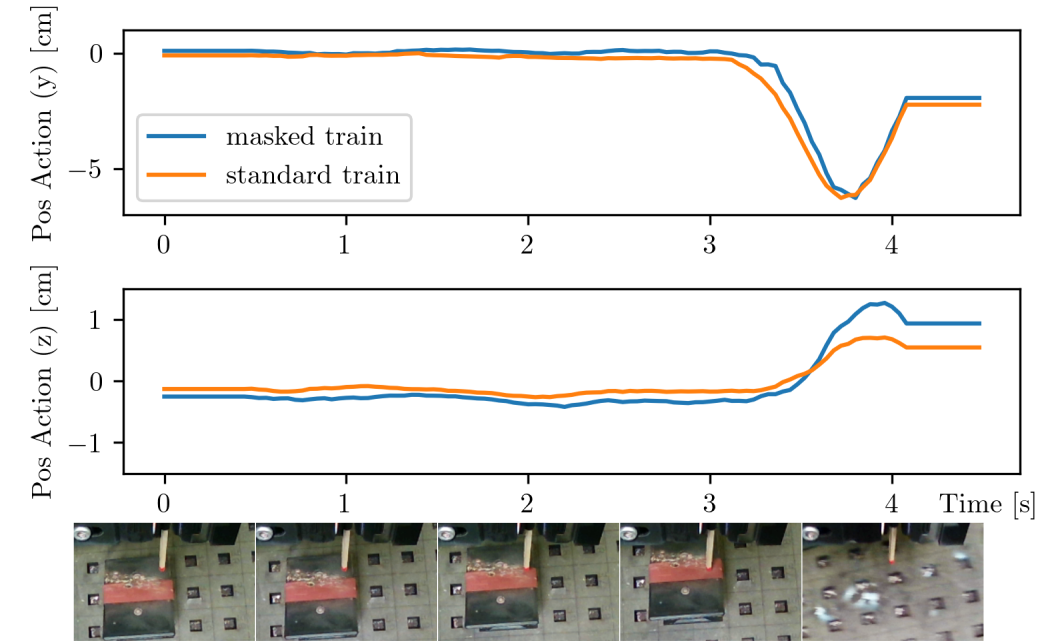}
\end{center}
\caption{Comparing the predictions of two vision-only policies (one trained with the standard procedure, the other one with the masked one). We visualize the policy predictions for the y- and z-component of the 5th action on a trajectory that was obtained by rolling out the standard policy and on which it fails to establish sufficient contact between the match and the striker paper. As shown, the policy that underwent the masked training procedure proposes different actions, i.e., moving closer to the striker paper before accelerating along the striker paper (as shown for the z-predictions when $T<\SI{3.5}{\s}$).
Additionally, it proposes to accelerate at a later point in time along the striker paper, as shown for the y-axis predictions.}
\label{fig:compare_ac_gen}
\vspace{-0.25cm}
\end{figure}

\begin{table}[t]
\centering
\caption{
Success Rate of different vision-only policies in the variable grasp scenario.
The policies differ regarding the training procedure, i.e., whether they are trained with the standard procedure or with masked training that considers the tactile signals during training.
The masked training procedure, i.e., leveraging touch during training, is effective and yields increased success rates.
}
\label{tab:comparison_vargrasp_masekd_train}
\scalebox{1.}{
\begin{tabular}{l|cc}
\multicolumn{1}{c|}{\multirow{2}{*}{}} & \multicolumn{2}{c}{Policy Configuration} \\
 & Vision & Vision (masked training) \\ \hline
Success Rate & 20\% (8\%) & \textbf{40\% (8\%)} \\ \hline
\end{tabular}
}
\vspace{-0.25cm}
\end{table}

As shown in Tab. \ref{tab:comparison_vargrasp_masekd_train}, the vision-only policies that have undergone the masked training procedure achieve higher success rates, increasing the number of successful rollouts by a factor of 2, achieving a 40\% success rate.
While policies trained with the standard procedure often fail to establish sufficient contact between the match and the striker paper (46\% of rollouts fail due to no contact or insufficient forces as shown in Fig.~\ref{fig:failure_analysis}), policies trained with the masked training procedure exhibit a lower probability of these failures (25\%).
To underline this finding quantitatively, Fig.~\ref{fig:compare_ac_gen} compares the differently trained policies regarding action generation.
It visualizes the translational outputs for the 5th action in the sequence along the y- (direction of acceleration along the striker paper) and the z-direction (controlling the height of the match tip).
For the comparison, we consider a trajectory that has been obtained by rolling out the policy trained with the standard procedure.
During this trajectory, the policy failed to establish contact between the match and the striker paper.
Considering the z-component of the predicted action, before the start of the sideways motion, the policy that was trained using the masked procedure outputs lower values, thereby indicating that it wants to move the end effector lower, increasing the probability of making contact with the striker paper.
Considering the y-direction, it is also evident that the policy trained using the standard procedure aims to move along the striker paper earlier.
This behaviour again increases the probability of accelerating too early without making proper contact with the striker paper.
We conclude that the masked training procedure increases the success rates of vision-only policies.
Therefore, the availability of tactile observations can improve policy performance, even when it is only provided during training.

\subsection{How does the performance difference evolve when increasing the number of available demonstrations?}

\begin{figure}
\begin{center}
\includegraphics[width=0.95\columnwidth]{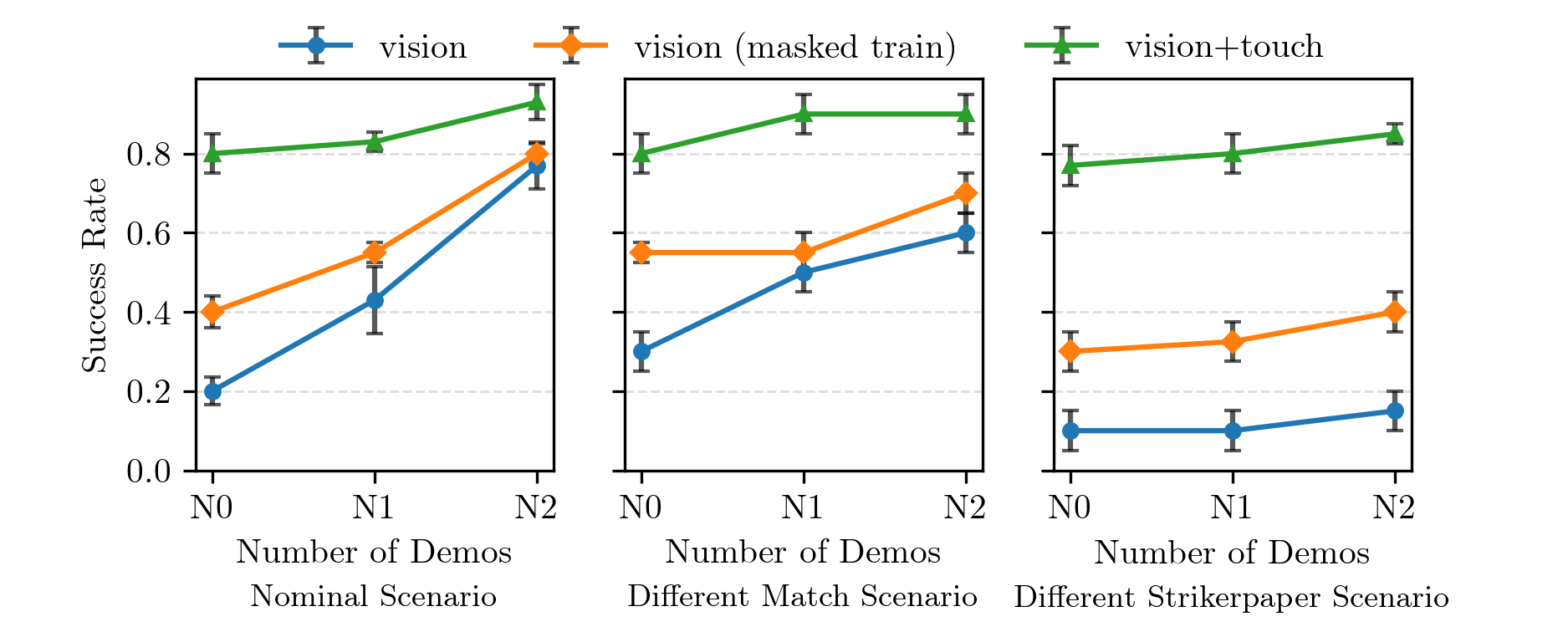}
\end{center}
\caption{Data scaling study. Comparing success rates across varying number of demonstrations, while considering the three different scenarios of the nominal configuration (left), considering different matches (middle), and a different striker paper (right). For training the policies, we consider the standard dataset N0 (20 demonstrations), N1 (45 demonstrations), and N2 (70 demonstrations).}
\label{fig:data_scaling}
\vspace{-0.45cm}
\end{figure}

This section evaluates the scalability of our findings by investigating policy performance as the number of demonstrations increases.
Specifically, we compare policies trained on $N0{=}20$, $N1{=}45$, and $N2{=}70$ demonstrations. This analysis seeks to determine whether vision-only policies can eventually bridge the performance gap relative to vision+touch policies as data availability increases. To further test the policies' robustness, we evaluate them in the nominal scenario and in two new environments that include using (1) thinner matches, and (2) a different striker paper (cf.~\fig{fig:initializations} - middle \& right).\\
The results are shown in Fig. \ref{fig:data_scaling}. We find that the vision+touch policies consistently outperform both the standard vision-only policies and those enhanced by masked training across all scenarios. While the performance of vision-only policies improves with more data, particularly in the nominal scenario, they only approach the baseline performance of the visuotactile policy (trained on $N0{=}20$) when provided with $N2{=}70$ demos. In other words, in the nominal scenario, the vision-only policies require more than three times the amount of data to eventually reach the performance of the multimodal (vision+touch) policies trained on the fewest demonstrations.\\
In both generalization scenarios, the performance disparity persists. While increasing $N$ helps vision-only policies extract better features for the nominal scenario, they struggle to build representations robust enough to handle task-relevant perturbations. This limitation is most evident when exchanging the striker paper.
Overall, these findings underline that actively incorporating the tactile signals into the visuomotor policies is essential not only for maximizing performance in low-data regimes but also for achieving robust generalization to novel evaluation scenarios, since the touch features allow for reliably extracting the contact-related features, which are critical for high task success rates and robust performance.

\subsection{How does our approach perform compared to baselines?}

\begin{table}[t]
\centering
\caption{
Success rates of different visuotactile policies on the match lighting task. The policies differ w.r.t. architectures and training objectives. Our proposed architecture, together with the flow-matching objective, yields the best performance.
}
\label{tab:comparison_baselines}
\scalebox{1.}{
\begin{tabular}{l|c}
Policy & Success Rate \\ \hline
See, Hear, and Feel \cite{li2023see} & 50\% (8\%) \\
Diffusion Policy (DP architecture + DDIM) \cite{chi2023diffusion} & 30\% (14\%) \\ 
Our architecture (III-B) + DDIM & 53\% (5\%) \\ 
\textbf{Ours} (Our architecture (III-B) + Flow (III-A)) & \textbf{80\% (8\%)} \\
\end{tabular}
}
\vspace{-0.25cm}
\end{table}

This section compares the performance of visuotactile policies trained using our proposed approach with baselines, using 20 human demonstrations.
As baselines, we consider See, Hear, Feel \cite{li2023see}, which similarly employs a multi-head attention-based policy architecture but differs in its training strategy: it relies on an explicit behaviour cloning loss, directly regressing to the action prediction.
We also compare with the vanilla implementation of Diffusion Policy (DP) \cite{chi2023diffusion} that uses the DDPM during training and DDIM sampling during inference.
To disentangle the effects of the policy architecture and the sampler, we create a third baseline that consists of our proposed policy architecture (cf. Sec. III-B) trained with DDPM, and using DDIM during inference.
To ensure a fair comparison, all policies are trained on the same vision+touch data, use identical observation encoder architectures, and are configured to have comparable parameter counts.
These choices ensure that all policies achieve real-time inference given our \SI{25}{Hz} control frequency.

The results in Tab.~\ref{tab:comparison_baselines} showcase that our proposed approach outperforms the baselines. The lower success rates of the See, Hear, and Feel~\cite{li2023see} baseline stem from difficulties in reliably reaching the striker paper, which we attribute to the behavioral cloning loss struggling with multi-modal data during the approach phase.
While the DP baseline employs a more expressive generative model, the vanilla version performs worse in this comparison, with only 30\% successes on average.
Notably, exchanging the DP architecture with our proposed model architecture (our architecture + DDIM) improves success rates to 53\%.
Compared to our proposed architecture, the DP architecture concatenates the encodings of the individual observation modalities without enforcing equal dimensionality within the policy network. We hypothesise that this design choice accounts for the observed performance gap, which may be particularly impactful given the low-demonstration regime.
Lastly, the results showcase another improvement in performance to 80\% upon combining our architecture with the proposed flow-matching generative model. 
Given the budget of only 5 inference iterations, the straight-line rectified flow produces more precise and less noisy actions compared to DDIM, which is crucial for success in robotic match lighting that requires both precision and reactivity.
To further support this choice of limiting inference to $K{=}5$ iterations, we conducted an ablation with $K{=}10$ for our approach. While the choice of $K{=}10$ increased the inference time to \SI{55}{ms}, performance dropped to a 50\% (10\%) success rate, due to the less regular action updates, which increase the likelihood of the policies accelerating before establishing sufficient contact.
Overall, these results underline the importance of our approaches' individual components and their competitive performance.

\subsection{Do the results transfer to the task of whiteboard erasing?}

\begin{figure}
\begin{center}
\includegraphics[width=\columnwidth]{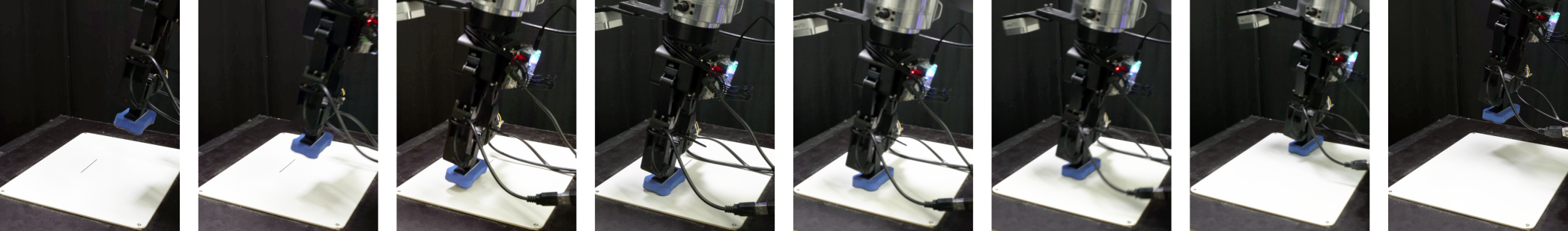}
\end{center}
\caption{
Illustrating task progression (from left to right) for the whiteboard erasing task. First, the eraser must be brought into contact with the whiteboard; then, the policy must move along the mark while ensuring sufficient contact force; finally, the eraser should be lifted.}
\label{fig:whiteboard_erasing}
\vspace{-0.25cm}
\end{figure}

\begin{table}[t]
\centering
\caption{
Policy success rates for the whiteboard erasing task. Apart from the vision-only policies, all models use both vision and touch observations. Our proposed approach achieves the best performance.
}
\label{tab:comparison_baselines_erase}
\scalebox{.9}{
\begin{tabular}{l|c}
Policy & Success Rate \\ \hline
Vision-only (Our architecture (III-B) + Flow (III-A)) & 13\% (19\%) \\
See, Hear, and Feel \cite{li2023see} & 33\% (19\%) \\
Diffusion Policy (DP architecture + DDIM) \cite{chi2023diffusion} & 27\% (9\%) \\ 
Our architecture (III-B) + DDIM & 40\% (16\%) \\ 
\textbf{Our Vision+Touch} (Our architecture (III-B) + Flow (III-A)) & \textbf{80\% (16\%)} \\
\end{tabular}
}
\vspace{-0.25cm}
\end{table}

This final experiment evaluates the generalizability of our findings to another manipulation task: erasing a 6~cm whiteboard marking (cf.~\fig{fig:whiteboard_erasing}). As with match lighting, this task requires establishing contact and maintaining a consistent force while moving along the board to erase the line. We collected 20 demonstrations through kinesthetic teaching and introduced task variability by grasping the eraser with translational and rotational perturbations of $\pm$1~cm \& $\pm$10$^{\circ}$, respectively.\\
The results in Table~\ref{tab:comparison_baselines_erase} underscore the effectiveness of our proposed approach. Consistent with previous evaluations, the vision-only baseline yields substantially lower success rates, confirming the importance of incorporating tactile feedback for this task as well.
Specifically, vision-only policies struggle to resolve the transition from approach to interaction, and often fail to maintain consistent normal forces between the eraser and the whiteboard required for effective erasure.
Furthermore, our approach outperforms the other baselines that leverage both vision and tactile signals. Our framework outperforms the See, Hear, and Feel~\cite{li2023see} baseline, which continues to struggle with the inherent variability of the demonstration data and with aligning the eraser with the markings.
The findings also further reaffirm that ensuring equal dimensionality between the individual modalities is important. Without this alignment, the performance is reduced, as observed in the performance difference between the DP baseline and the baseline using our architecture and the DDIM sampler.
Finally, the use of a flow-matching generative model remains a second key driver of success.
The gains are particularly pronounced in this task, likely because successful execution requires precise, sustained action generation over a longer duration as the robot traverses the marking.
Overall, these results demonstrate that our framework generalizes effectively to other contact-rich tasks while maintaining high performance in low-demonstration regimes.

\subsection{Discussion}

Our experimental evaluation focused on a high-dimensional optical tactile sensor and a parallel gripper to investigate the role of touch sensing in dynamic manipulation tasks with a particular focus on robotic match lighting. As supported by our results and the attention analysis, we believe these findings underscore a general principle of contact observability: policies benefit from direct contact observations that provide informative features crucial for understanding and controlling the contact configuration, thereby reducing contact-related failures and improving performance. We hypothesize that these benefits transfer to other tactile sensors and end-effectors.
However, future benchmarking is required to investigate and isolate how specific sensor properties, placements, and end-effector geometries influence policy learning. Additionally, investigating how policies can be further improved by, for example, learning from unsuccessful demonstrations, or incorporating even higher frequency tactile feedback as in~\cite{xue2025reactive, chen2025implicitrdp}, remains a promising direction for future work.

\section{CONCLUSION}
\label{sec:conclusion}

	This work investigated the importance of incorporating touch sensing to learn dynamic manipulation tasks from a few human demonstrations, with a particular focus on robotic match lighting.
We also introduced a policy learning framework that combines a flow-matching generative model for fast, efficient action generation with an expressive and modular transformer architecture.
The experimental results showcased the effectiveness of our proposed approach compared with competitive baselines.
They also showed that the performance benefits extend beyond match lighting and hold for a whiteboard-erasing task.
Moreover, our thorough experimental evaluation highlighted the importance of incorporating touch sensing into the visuomotor policies, which reliably improved performance across all tasks and data regimes.
We demonstrated that integrating tactile signals effectively reduces contact-related failures, thereby improving success rates and enhancing the policies' robustness and generalization.
Taken together, these findings highlight the synergistic potential of integrating tactile sensing with suitable policy architectures to learn performant policies for dynamic manipulation.

\section*{ACKNOWLEDGMENT}
We thank Erik Helmut and Rickmer Krohn for helping with the 3D printing and attention analysis.
This work was supported by the German Research Foundation (DFG) Emmy Noether Programme (CH 2676/1-1), the EU’s Horizon Europe project ARISE (Grant no.: 101135959), the German Research Foundation (DFG) under the Cluster of Excellence CARE: Climate-Neutral And Resource-Efficient Construction (EXC 3115) project number 533767731, by the project "Genius Robot" (01IS24083), funded by the Federal Ministry of Education and Research (BMBF), and by Bundesministerium f\"ur Bildung und Forschung (BMBF) and German Academic Exchange Service (DAAD) in project 57616814 SECAI.

\bibliographystyle{IEEEtran}
\bibliography{bibliography}

\newpage
\section*{APPENDIX}
\subsection{Additional details regarding policy inference \& control}

We use the Cartesian Impedance Controller from \cite{Nbfigueroafranka_interactive_controllers} to move the robot during the autonomous policy rollouts.
We tuned the controller's stiffness and damping values on a few of the collected demonstration trajectories.
The gains have been chosen such that replaying the trajectories obtained during kinesthetic teaching yields task success when tracked using this control strategy.
We rely on the Robotic Operating System (ROS) to gather the sensor observations.
Policy inference is run asynchronously, and only the first action of the action sequence is applied by the controller before updating the action sequence based on the most recent model inference.
The policies run online in real-time as action generation, i.e., policy inference, only takes \SI{0.028}{\s} for our largest vision+touch policies on an NVIDIA $3090$ GPU.

\subsection{Additional insights regarding the importance of tactile feedback for obtaining performant match lighting policies under fixed grasping poses}

\begin{figure}
\begin{center}
\includegraphics[width=0.95\columnwidth]{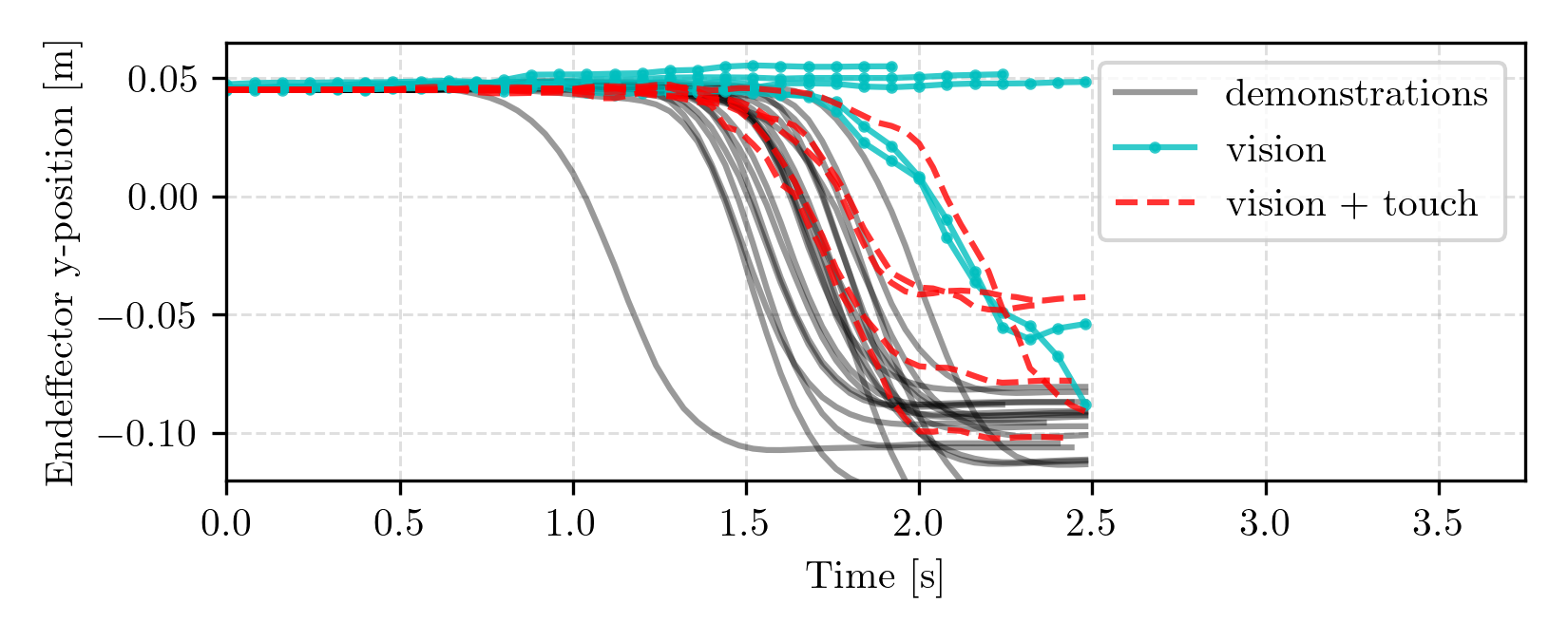}
\end{center}
\caption{Comparing the demonstrated trajectories with trajectories from rolling out different policies, considering the y-position in meters of the end effector.
The y-coordinate is the direction along the striker paper in which the robot needs to accelerate to light up the matches.
Qualitatively, the vision+touch policies generate rollouts that better match the demonstrations compared to the vision-only policies, indicating that the tactile observations contain important information for explaining and matching the demonstrations.
}
\label{fig:task_success_rate}
\end{figure}

\begin{figure}
\begin{center}
\includegraphics[width=0.4\columnwidth]{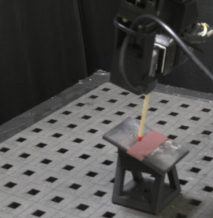}
\end{center}
\caption{Visualizing the initial configuration during the fixed grasping experiment.
}
\label{fig:initializations_new}
\end{figure}

Contrary to the findings presented in Section IV-A of the main paper, this section presents an additional experiment that investigates the importance of tactile sensing for obtaining effective match-lighting policies, now considering fixed grasp poses.
In particular, to conduct this experiment, we collected 20 additional demonstrations using a fixed grasping pose, as visualized in Fig.~\ref{fig:initializations_new}.

In this fixed grasp pose scenario, we find that the vision+touch policies outperform the vision-only policies, achieving a mean success rate of 87\% compared to 33\%, with both having a standard deviation of 12\%. 
Apart from the differences in success rate, Fig.~\ref{fig:task_success_rate} reveals that the rollouts of the vision+touch (also referred to as visuotactile) policies better match the demonstration data.
The visuotactile policy evaluations better align in terms of the timing of accelerating along the striker paper, which corresponds to the end-effectors y-axis.
This finding hints that vision-only policies struggle to precisely detect the point in time of making contact since this indicates that the acceleration phase along the striker paper should follow, and therefore underlines the importance of incorporating tactile sensing to obtain trajectories during policy inference that better match the human demonstrations.

\subsection{Evaluating the robustness of the visuotactile policies robust w.r.t. generalizing to novel scenarios}

\begin{table}[t]
\centering
\caption{
Success Rate of our vision+touch policies when evaluating the policies in novel, previously unseen scenarios.
}
\label{tab:generalization_exps}
\scalebox{.95}{
\begin{tabular}{l|c}
Evaluation Configuration & Success Rate \\ \hline
Different Mounting Angle (5$^{\circ}$) & 77\% (5\%) \\
Different Mounting Angle (30$^{\circ}$) & 67\% (5\%) \\ 
Nominal Mounting Angle (20$^{\circ}$), Different Striker Paper & 77\% (9\%) \\
Actual Matchbox (handheld) \& Different Striker Paper & 70\% (8\%) \\
Varying Lighting Conditions & 67\% (5\%) \\
\end{tabular}
}
\end{table}

\begin{figure}
\begin{center}
\includegraphics[width=0.99\columnwidth]{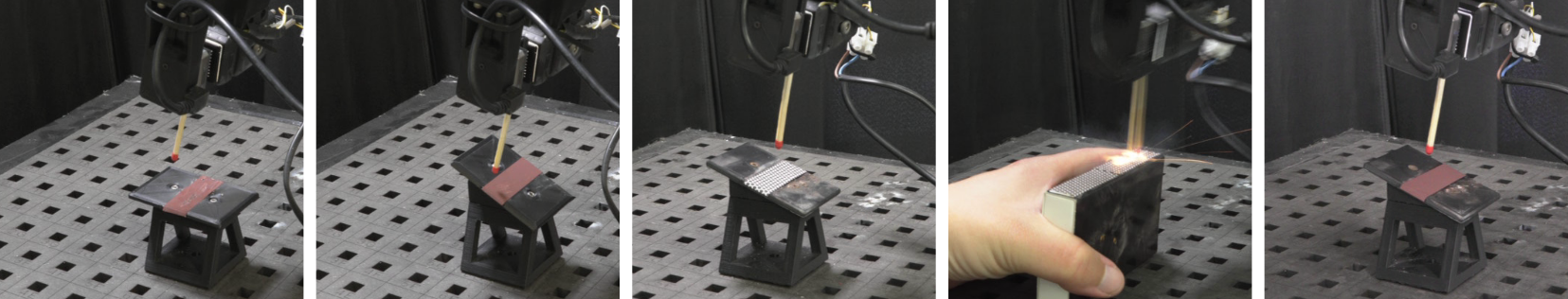}
\end{center}
\caption{Visualizing the experiment setups considered in the generalization experiments. From left to right: Mounting angle of 5$^{\circ}$; Mounting angle of 30$^{\circ}$; Nominal mounting angle of 20$^{\circ}$ but different striker paper; Using an actual handheld matchbox; Varying lighting conditions (decreased illumination). Note how the different mounting angles alter the angle \& distance between match \& striker paper, and how the other changes affect the visual appearance.
}
\label{fig:mounting_angles}
\end{figure}

This additional experiment evaluates whether the visuotactile policies can generalize to novel, previously unseen scenarios (cf. Fig.~\ref{fig:mounting_angles}).
We evaluate the following variations: (1) altering the angle between the match and the striker paper by mounting the paper at previously unseen angles of 5$^{\circ}$ and 30$^{\circ}$; (2) using a different, dotted striker paper; (3) replacing the 3D-printed mount with an actual matchbox that is handheld; and (4) varying lighting conditions by increasing or decreasing the intensity of the external light source.
This evaluation considers the visuotactile policy with the pretrained+refine training procedure and the variable grasp initialization.

In addition to providing rollout videos in the supplementary material, quantitatively, as shown in Tab.~\ref{tab:generalization_exps}, the policies generalize well to the 5$^{\circ}$ mounting angle and to the different striker paper, with only a 3\% drop in mean success rate compared to the policies' mean success rate of 80\% in the nominal scenario. In contrast, performance decreases by 10\%, 13\%, and 13\% for the handheld matchbox, 30$^{\circ}$ mounting angle, and varying lighting conditions, respectively.
The slightly lower successes at 30$^{\circ}$ can be attributed to the match starting closer to the 3D-printed holder, leaving less room for adjusting the angle relative to the striker paper. We further hypothesize that the handheld matchbox and lighting variations introduce the most substantial visual perturbations, leading to reduced successes as the policies more often fail to align the match tip with the striker paper in the initial phase.
Despite being trained on only 20 demonstrations, the results suggest that the learned visuotactile policies exhibit robustness for variations beyond the training scenario, generalizing reasonably well to new conditions, with the overall performance remaining above the performance of the previously investigated baselines in the nominal scenario. Future improvements could be achieved by adding demonstrations for the more challenging scenarios and by refining visual data augmentation strategies.

\subsection{Additional Information regarding the Statistical Significance of the Experimental Results}

This section provides a formal analysis of the statistical significance of the success rates reported in Sections IV-C and IV-D.\\
We employed Fisher’s Exact Test to evaluate the significance of the observed performance differences. This test was selected to account for the categorical (binary) nature of the success metrics and the relatively small sample size ($N=30$ per group). While all policies were evaluated under identical task distributions and hardware, factors such as stochasticity in the grasping pose, the precise placement of the striker paper, and physical wear introduced inherent randomness. Therefore, an independent exact test (i.e., the two-sided Fisher's Exact Test) was preferred over a paired test (such as McNemar’s) to avoid assuming a 1-to-1 correspondence between specific environmental perturbations.\\
Regarding the experiments in Section IV-C, considering the nominal scenario, the visuotactile (vision+touch) policies significantly outperformed both vision-only baselines (two-sided Fisher’s exact test, $p < 0.05$) for the low-data regimes ($N0$ and $N1$). Under the largest demonstration regime ($N2$), the performance gap narrowed: the differences were no longer significant, yielding a $p$-value of $0.14$ ($OR = 4.26$) compared to the vision baseline and a $p$-value of $0.25$ ($OR = 3.5$) compared to the vision (masked) baseline.\\
When evaluating generalization to a different match type, the visuotactile policies maintained significant superiority over vision-only baselines ($p < 0.05$), with one exception. Under the $N_2$ regime, the comparison with the vision-masked baseline yielded $p = 0.1$ ($OR = 3.86$), falling just outside the threshold for significance.\\
For the generalization experiments involving a different striker paper, the performance improvements provided by the visuotactile policies remained statistically significant across all demonstration regimes (two-sided Fisher’s exact test, $p < 0.05$).\\
For the experiments presented in Section IV-D, and in particular the baseline comparisons presented within Table II, the performance differences between our proposed approach (our architecture + flow) and the baselines are statistically significant (two-sided Fisher’s exact test, $p < 0.05$) apart from the comparison with the policy consisting of our architecture + the DDIM sampler.
This comparison against the baseline of (our architecture + DDIM) fell just outside the standard significance threshold ($p = 0.0539$), the associated odds ratio of 3.5 still indicates a substantial relative improvement in success probability.

\end{document}